\documentclass[letterpaper, 10 pt, conference]{ieeeconf}  % Comment this line out if you need a4paper

\IEEEoverridecommandlockouts                              % This command is only needed if 
\usepackage{graphicx}
\usepackage{amsmath}
\usepackage{amssymb}
\usepackage{url}
\usepackage{subcaption}
\usepackage{float}
\usepackage{multirow}
\usepackage{booktabs}
\usepackage{hyperref}
\usepackage{amsmath}
\title{\LARGE \bf
ALDM-Grasping: Diffusion-aided Zero-Shot Sim-to-Real Transfer for Robot Grasping}

\author{Yiwei Li\textsuperscript{*}$^{1}$, Zihao Wu\textsuperscript{*}$^{1}$, Huaqin Zhao$^{1}$, Tianze Yang$^{1}$, Zhengliang Liu$^{1}$, Peng Shu$^{1}$, 
\\Jin Sun$\dagger$$^{1}$, Ramviyas Parasuraman$\dagger$$^{1}$ and Tianming Liu$\dagger$$^{1}$
\thanks{\textsuperscript{*}Equal Contribution.}%
\thanks{$\dagger$Corresponding authors: Tianming Liu (tliu@uga.edu), Ramviyas Parasuraman (ramviyas@uga.edu), and Jin Sun (jinsun@uga.edu)}
\thanks{$^{1}$School of Computing, University of Georgia, Athens, GA 30602, USA}%}
}

\begin{document}

\maketitle
\thispagestyle{empty}
\pagestyle{empty}

%%%%%%%%%%%%%%%%%%%%%%%%%%%%%%%%%%%%%%%%%%%%%%%%%%%%%%%%%%%%%%%%%%%%%%%%%%%%%%%%
\begin{abstract} To tackle the "reality gap" encountered in Sim-to-Real transfer, this study proposes a diffusion-based framework that minimizes inconsistencies in grasping actions between the simulation settings and realistic environments. The process begins by training an adversarial supervision layout-to-image diffusion model(ALDM). Then, leverage the ALDM approach to enhance the simulation environment, rendering it with photorealistic fidelity, thereby optimizing robotic grasp task training.  Experimental results indicate this framework outperforms existing models in both success rates and adaptability to new environments through improvements in the accuracy and reliability of visual grasping actions under a variety of conditions. Specifically, it achieves a 75\% success rate in grasping tasks under plain backgrounds and maintains a 65\% success rate in more complex scenarios. This performance demonstrates this framework excels at generating controlled image content based on text descriptions, identifying object grasp points, and demonstrating zero-shot learning in complex, unseen scenarios.

% The advent of deep learning has significantly reshaped the landscape of robotic downstream tasks, offering innovative pathways to mitigate the limitations posed by the dearth of annotated training data through Sim-to-Real transfer methodologies. However, the persistent challenge of the "reality gap"—the divergence between conditions in simulated training environments and those in actual operational contexts—continues to impede the seamless translation of these models into practical applications. This disparity constitutes a formidable barrier, as it often leads to suboptimal performance of models, originally trained in simulations, upon their transition to real-world environments.
\end{abstract}

%%%%%%%%%%%%%%%%%%%%%%%%%%%%%%%%%%%%%%%%%%%%%%%%%%%%%%%%%%%%%%%%%%%%%%%%%%%%%%%%
\section{INTRODUCTION}

% Deep learning has significantly impacted robotics, enhancing its capabilities and applications. 
To perform optimally, deep learning models require extensively annotated datasets~\cite{karoly2020deep}. However, in robotics tasks such as visual grasping, acquiring datasets to train comprehensive deep learning models in real-world settings can be prohibitively expensive and time-consuming, and sometimes, unfeasible~\cite{song2020grasping,kalashnikov2018scalable,bousmalis2018using}. This problem is also known as the "reality gap". To address this challenge, Sim-to-Real strategies have been developed, allowing the training of robotic performance models, such as those for visual grasping, to be trained in simulated environments and subsequently transferred to real settings through techniques like domain randomization and domain adaptation~\cite{bousmalis2018using,james2019sim}. Domain randomization diversifies visual elements (e.g., textures, colors) in training simulations to make models focus on invariant aspects of images applicable in real-world scenarios. Conversely, domain adaptation tailors models from the simulated (source) domain for the real (target) domain, often leveraging unlabeled target domain data~\cite{tobin2018domain}. 

\begin{figure}[ht!]
	\centering \tiny
% 	\vspace*{-0.2cm}
	\begin{subfigure}[b]{\linewidth}  
		\centering 
		\includegraphics[width=1.0\textwidth]{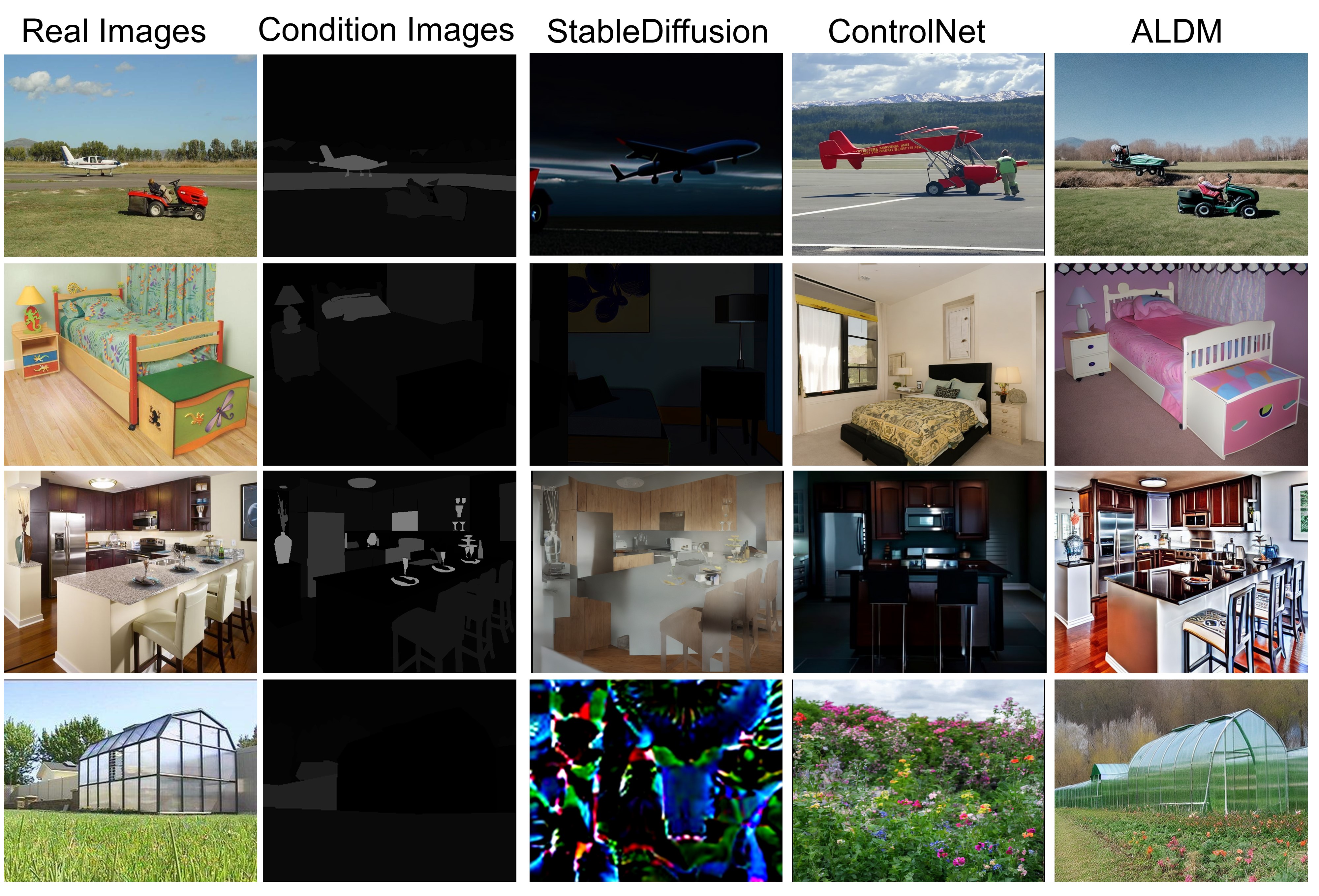}%\vspace*{-0.2cm}
		\caption[]%
		{{Generated image results of different models trained in ADE20k dataset.
		}}    
		\label{fig:Spatial-data}
	\end{subfigure}
	\hfill
	\begin{subfigure}[b]{0.48\textwidth}  
		\centering 
            % \hspace*{-1.0cm}
		\includegraphics[width=\textwidth]{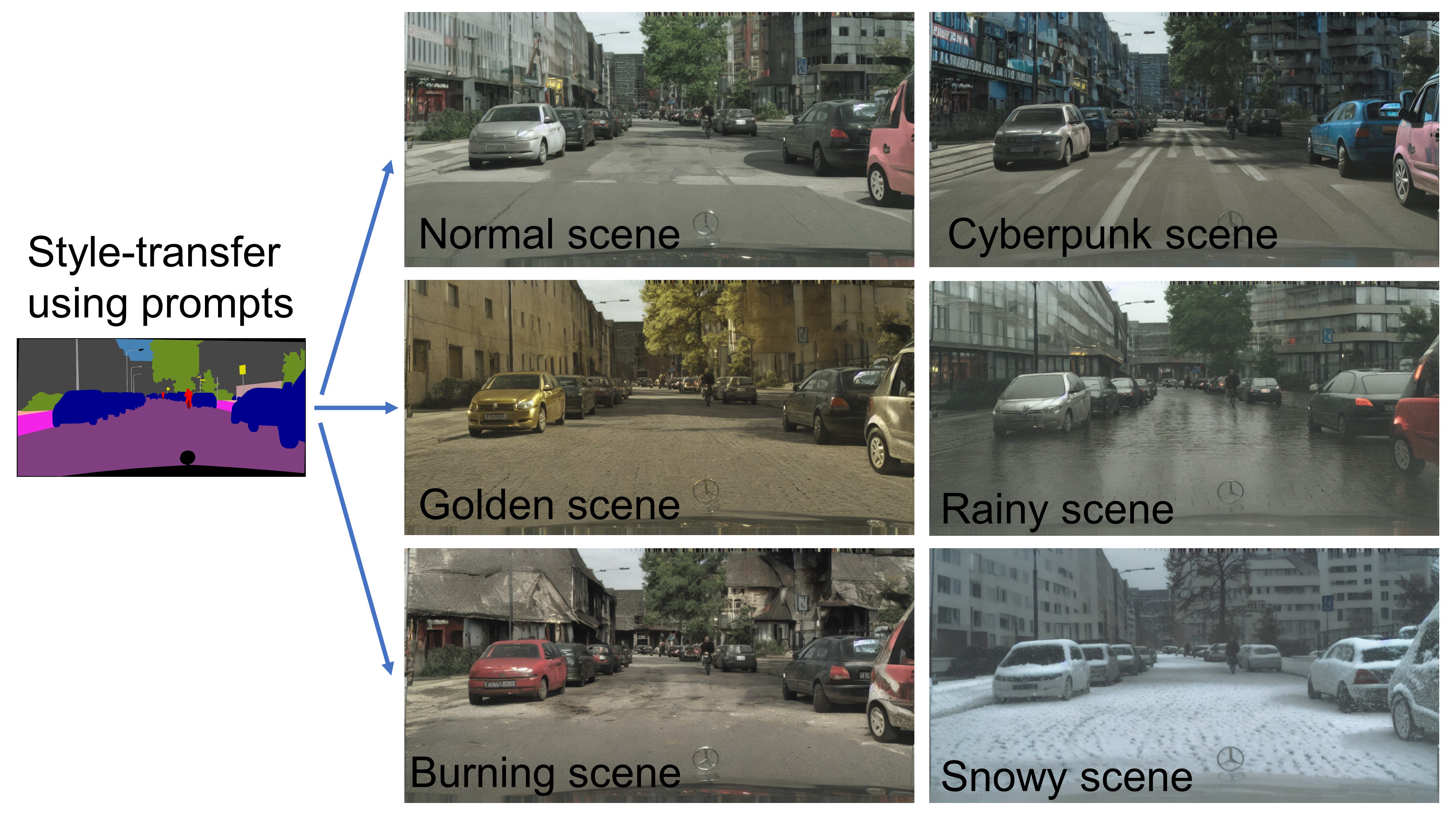}%\vspace*{-0.2cm}
		\caption[]%
		{{Generated image results of ALDM trained in Cityscapes dataset.
		}}    
		\label{fig:temporal}
	\end{subfigure}
	\hfill
	\caption{{
	ALDM has great performance in zero-shot image generation and style transfer. The image generation model used in this article contains data of various types and can generate objects in various robotic application scenarios. The generated image is not only close to the realistic-style scene but also ensures that the number and position of key objects are consistent with the original image. There is huge potential to provide more diverse, realistic, and accurate training samples for robot action planning applications.}
	} 
	\label{ALDM_examples}
 \vspace{-4mm}
\end{figure}

In image transfer tasks, GANs, such as RetinaGAN~\cite{ho2021retinagan}, CycleGAN~\cite{zhu2017unpaired} are usually used as the main technical framework~\cite{liu2021review}. Some variants, such as DT-CycleGAN~\cite{liu2023digital} and StarGAN~\cite{choi2018stargan}, are also invented to address different challenges. A major advantage of these kinds of models is that their training process does not require data to appear in pairs, which allows them to show better performance on specific tasks. However, GAN models also have some limitations, the most significant of which are their preliminary requirements for large amounts of training data and limitations in the model's generalization capabilities. The latter means that when faced with new tasks or scenarios, existing GAN models often need to be re-trained from scratch, which greatly constrains their applicability and flexibility in different application cases. In contrast, the advantages of diffusion models in this aspect are revealed. To effectively control the image effects generated by the diffusion model, the research field has developed a variety of technical paths, including layout-to-image, text-to-image, and image-to-image generations. 

\begin{figure*}
\centering
  \includegraphics[width=\linewidth]{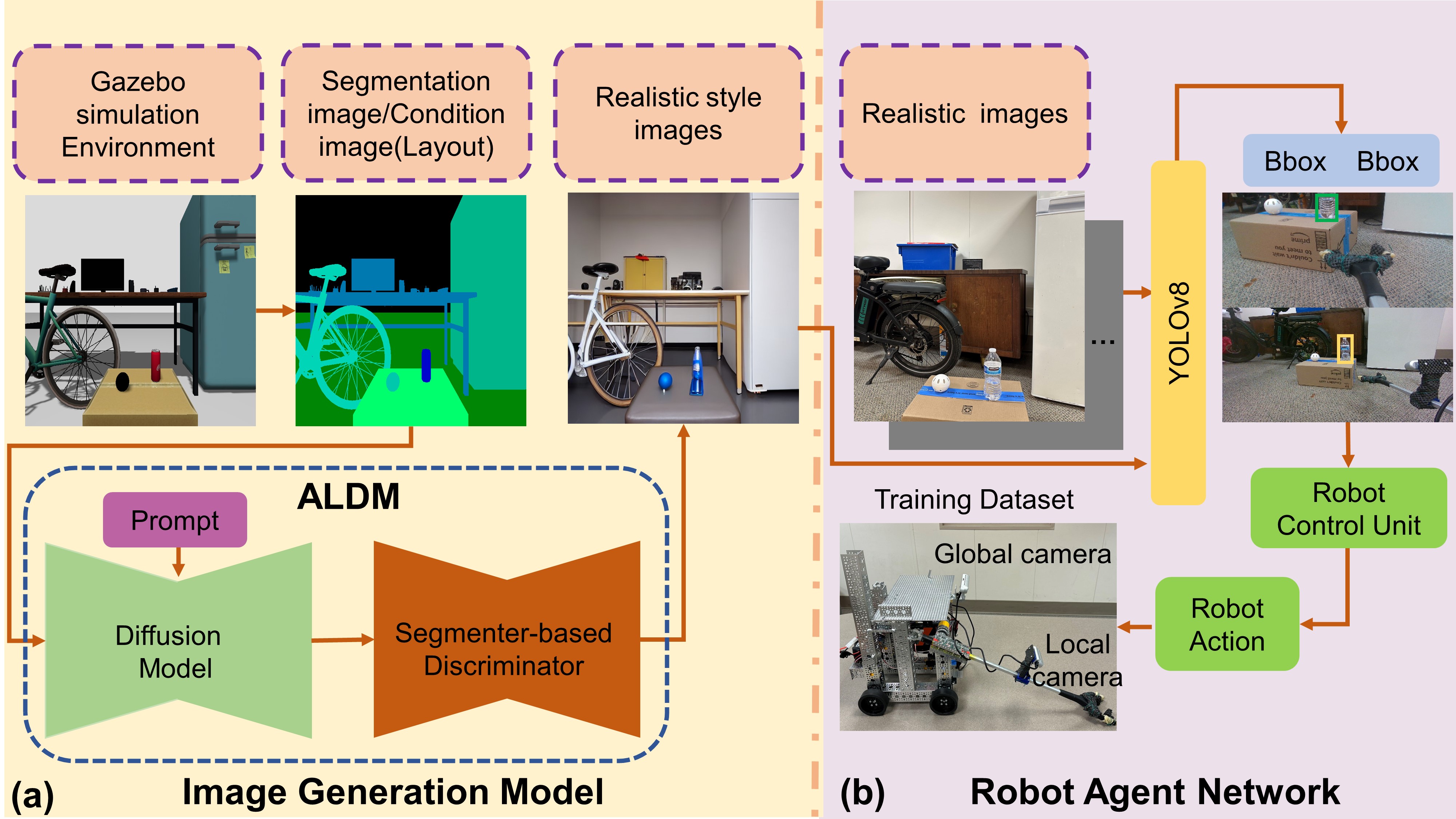}
\caption{The whole pipeline of the diffusion-model-based grasping robot. a) The generation procedures for realistic style images. b) The robot agent network designed for the task of object detection and grasping.}
\label{fig:pipeline}       % Give a unique label
\end{figure*}
ControlNet~\cite{zhang2023adding}, an innovative deep learning framework, is engineered to exert meticulous control over generative models' outputs, with a particular emphasis on image synthesis tasks. This sophisticated architecture exhibits exceptional abilities in transfer-learning from prompts to images, transcending the conventional constraints paradigms. ControlNet is adept at interpreting textual prompts, thereby furnishing users with the ability to steer the image generation process with accurate content of the images and adaptability. The core idea usually involves guiding the generation process through conditional input, thereby achieving fine control over the properties of the generated image (such as layout, style, content, etc.). This method is particularly suitable for application scenarios that require highly customized output. However, although ControlNet can utilize a condition image as input to constrain the output the training of ControlNet is more complicated, and the training results are relatively uncontrollable. Although there is a condition image as input, it may still cause deviations in the positions, sizes, numbers, and arrangements of major objects in the output, so the ControlNet is not suitable for this task. Fine-tuning ControlNet is a potential solution, but the fine-tuning process requires a large amount of data. It does not necessarily guarantee results, which means it is not an efficient and reliable solution.

One possible way to solve the deviations is to use the layout-to-image method, which aims to generate realistic images based on predetermined layout information~\cite{zheng2023layoutdiffusion}. In this framework, "layout" generally refers to an abstract representation of objects in an image, covering their categories, locations, sizes, and their interrelationships. This method offers a significant advantage in robotic grasping by allowing precise manipulation of both the layout and content within generated images. This precision is particularly beneficial for the transfer of simulated images to robot training, as maintaining consistent positioning and relative relationships among target objects is critical for avoiding later object detection training failures. However, this scheme also has a limitation that cannot be ignored, which is that labels of different layouts are the only control over the content of images, which makes the controls on the image contents simple and general.

So in comparison, the adversarial supervised layout-to-image diffusion model (ALDM)~\cite{li2024adversarial} is especially suitable for the Sim-to-Real task~\cite{li2024adversarial}. 
In the training phase of the ALDM, a segmenter-based discriminator is connected in series with a diffusion model. This configuration facilitates the quantitative assessment of the discrepancy between the segmented outcomes of images produced by the diffusion process and the ground-truth segmentation data. Consequently, the resulting model generates images where both the objects and spatial arrangement closely approximate reality (Fig. \ref{ALDM_examples}(a)). Simultaneously, the model preserves the prompt control feature inherent in diffusion models. This characteristic ensures that the synthesized images not only align well with the Sim-to-Real (simulation to reality) objectives pertinent to robotic applications but also support the facile adaptation to diverse experimental styles (Fig. \ref{ALDM_examples}(b)). Hence, the images crafted by this model are exceptionally conducive to robotic training across varied scenarios. The main contributions of this work are as follows:
\begin{itemize}
    \item We build a task learning pipeline on top of the state-of-the-art ALDM approach to the task of the Sim-to-Real robotic training and use quantitative results on generated images' segmentation to show that this method is highly appropriate for the Sim-to-Real robotic grasping task.
    \item We design a pipeline to train and control a two-camera robot to work in reality-unseen environments. The results show that the robot performs well in grasping unseen objects in unpredictably complex circumstances, which proves the reliability of the pipeline design.  
\end{itemize}

\section{RELATED WORK}
% \subsection{General Grasping and Visual Grasping}
% \subsection{Visual Grasping in Complex Background}
\subsection{Visual Grasping}

Visual grasping is an interdisciplinary field at the intersection of robotics, computer vision, artificial intelligence, and cognitive science, aiming to enable machines to interact with their environment through the detection, analysis, and manipulation of objects. This complex challenge involves both visual data processing and the control of robotic actuators~\cite{bicchi2000robotic,marwan2021comprehensive}. With advancements in deep learning, vision-based robotic grasping has become a focal point of research, adopting two primary methodologies: object reconstruction and end-to-end approaches~\cite{kleeberger2020survey}.

Object reconstruction methods focus on creating a detailed 3D model of the environment for precise planning and execution of grasps, offering accuracy at the cost of computational intensity and slower response times~\cite{bohg2013data}. Conversely, end-to-end methods leverage machine learning to directly map visual input to grasping actions without intermediate modeling, facilitating rapid adaptation to diverse scenarios albeit with less interpretability and higher demands on training data and computational resources~\cite{levine2018learning,song2020grasping,kalashnikov2018scalable,bousmalis2018using,kleeberger2020survey}.

\subsection{Image Generation Models for Bridging Reality Gap}
\subsubsection{GANs Models}
GANs revolutionized unsupervised learning by introducing a framework composed of a Generator, creating data that mimics real-world distributions, and a Discriminator, distinguishing between genuine and generated data. This innovation enables the production of realistic images, audio, and videos, significantly advancing the field of computer vision. Extending GANs, Zhu et al. proposed CycleGAN for unpaired image-to-image translation, utilizing a cycle consistency loss alongside adversarial losses to facilitate high-quality translations across different datasets without requiring paired examples. Building upon these advancements, Ho et al. introduced RetinaGAN~\cite{ho2021retinagan}, an object-aware sim-to-real adaptation technique that leverages object detection consistency to enhance the realism of translated images further. RetinaGAN's innovation lies in its ability to preserve object structures and textures across translations, offering significant improvements in sim-to-real tasks for robotics and demonstrating the adaptability of GANs to diverse visual environments.

\subsubsection{Diffusion Models}
% While Generative Adversarial Networks (GANs) laid the groundwork for generative modeling by simulating data distributions through adversarial training, diffusion models~\cite{rombach2022high} have proposed a new framework. They offer a paradigm shift by iteratively refining noise into complex images, circumventing some of the training stability and mode collapse issues associated with GANs. This breakthrough is significantly enhanced by contributions like ControlNet~\cite{zhang2023adding} and FreestyleNet~\cite{xue2023freestyle}. ControlNet innovatively introduces spatial conditioning to pre-trained text-to-image diffusion models, utilizing the structure of models such as Stable Diffusion to enable precise control over generated imagery based on specific spatial inputs like edges and poses. This allows for a tailored generation process with highly detailed and customized outputs. Concurrently, FreestyleNet advances the capabilities of image synthesis by combining layout and textual inputs in what is known as Freestyle Layout-to-Image Synthesis (FLIS). Using Rectified Cross-Attention (RCA), it adeptly marries textual semantics with spatial layouts, offering novel avenues for creative expression in generative imaging. These advancements underscore the versatility and adaptability of diffusion models, expanding the horizons of generative art and technology.
GANs initiated the era of generative modeling, but diffusion models represent a significant paradigm shift by transforming noise into detailed images through iterative refinement, addressing GANs' issues like training instability and mode collapse~\cite{rombach2022high}. Among these advancements, ControlNet~\cite{zhang2023adding} introduces spatial conditioning to text-to-image diffusion models, allowing for precise image generation based on specific inputs such as edges and poses. Concurrently, FreestyleNet~\cite{xue2023freestyle} extends image synthesis capabilities through Freestyle Layout-to-Image Synthesis (FLIS), which integrates textual and layout inputs using Rectified Cross-Attention (RCA). These innovations highlight the flexibility and potential of diffusion models in pushing the boundaries of generative art and technology.

\subsection{Robotic Application}
Sim-to-Real transfer is a crucial technique in robotics, aimed at bridging the gap between simulations and real-world applications~\cite{kleeberger2020survey,bousmalis2017unsupervised,peng2018sim,patel2015visual}. Key advancements include RetinaGAN, which improves simulated imagery for visual grasping through object-detection consistency~\cite{ho2021retinagan}, and RL-CycleGAN, which integrates Q-learning's Q-value with a reinforcement learning (RL) scene consistency loss to enhance Sim-to-Real transfer in RL applications~\cite{rao2020rl,watkins1989learning}. These GAN-based methods have significantly advanced the efficacy of visual grasping models by optimizing the use of virtual data.

In summary, our research integrates diffusion models with adversarial training to refine Sim-to-Real transfer~\cite{wang2022diffusion}. This hybrid approach combines the gradient-guided progression from noise to realistic data of diffusion models with the robust learning of adversarial methods, offering precise object locations, controllable image content generation, and zero-shot capacity for unseen scenario in Sim-to-Real applications. 
% This methodology leverages the strengths of both diffusion and adversarial training, yielding a framework that significantly benefits from the precision and adaptability in generating realistic imagery for robotic applications.

\section{METHODS}

\subsection{Robot Training and Robotic Control Pipeline}
To train a robot to grasp, the whole working pipeline can be seen in Figure \ref{fig:pipeline}. The whole procedure contains two parts:
\begin{itemize}
    \item \textbf{Image generation model} part firstly builds the simulation environments in the gazebo simulation platform. The labels of different objects can be set into the gazebo, thus the segmentation image can be easily obtained. These segmentation images are fed into the diffusion model as the condition images/layout images, and the ground truth labels are used for the training of object detection. In this part (Figure \ref{fig:pipeline}(a)), segmentation images are set as the layout and passed with the prompt to the ALDM, which is a U-Net~\cite{ronneberger2015u} structure. The generated images, produced by the diffusion model, will be passed to a pre-trained segmenter-based discriminator. The discriminator will compare the segmentation results of the generated image and the layout image to guarantee the content of the generated realistic-style image consistent with the input. 
    \item \textbf{Grasping Agent Network}
    After the realistic-style images are generated based on the simulation environments, these generated images will be used as the training data for the object detection model (YOLOv8 \cite{reis2023real}). The segmentation images will be treated as labels. The two RealSense RGB-D cameras on the physical robot will pass the image data to the pre-trained YOLOv8 model to separately generate the bounding box. Finally, the Control Unit will calculate the distance between bounding boxes to decide the robot base and gripper movement.
\end{itemize}

\subsection{Layout-to-Image Diffusion Model}
ALDM innovates the field of image synthesis by introducing adversarial supervision within the training framework of diffusion models. The key to this advancement is the employment of a discriminator, a segmentation-based model, that furnishes explicit, pixel-level feedback on the alignment between the generated images and the input layout. This mechanism not only bolsters the fidelity of the generated images to the specified layout but also preserves the versatility of text-driven editability. Such advancements are important for applications demanding high precision in generated content, such as robotic grasping, where accurate object positioning and count are paramount for effective training. 

The segmenter functions as a discriminator, tasked with classifying per-pixel class labels in authentic images against corresponding ground-truth maps. Simultaneously, it designates the synthetic outputs from the diffusion model as an additional "fake" category \(l\). Suppose the input(layout image) of the diffusion model is \(I_{image}\) and the output(generated image) is \(\hat{I}_{image}\). Given that the discriminator's role is fundamentally to address a multi-class semantic segmentation challenge, its training objective emanates from a foundational cross-entropy loss:

\begin{equation}
\begin{split}
    L_{Dis} = -\mathbb{E} \left[
    \sum_{c=1}^{N} \gamma_c \sum_{i,j}^{H \times W} l_{i,j,c} \log(Dis(I_{image})_{i,j,c})] \right] \\
     - \mathbb{E}\left[\sum_{i,j}^{H \times W}\log(Dis(\hat{I}_{image})_{i,j,c=N+1})\right],
\end{split}
\end{equation}

where \(N\) is the number of the semantic categories with \(H \times W\) represents the spatial size of the input. To control the importance of different objects 

\begin{equation}
   \gamma_c = \frac{H \times W}{\sum \mathbb{E}[l_{i,j,c} = 1]}
\end{equation}

\subsection{Object Detection and Grasping}

To fulfill the object detection, the grasping agent framework utilizes dual-camera imagery as its primary input to deduce the precise location for object grasping. This system incorporates two distinct visual perspectives: one providing a broad overview from the robot's base, termed the global view, and another offering a detailed focus on the gripper arm, referred to as the local view. The global view guides the movement of the robot base and the local view is responsible for adjusting the position of the gripper. The two cooperate to achieve accurate grasping of objects.  Thus, the system's input is symbolized as \(x\) in the domain of \(x \in \mathbb{R}^{B2{\times}C{\times}H{\times}W}\), integrates these two visual feeds, where \(B2\) signifies the batch size coupled with the dual imagery inputs, and \(C{\times}H{\times}W\) delineates the images' channel (C), height (H), and width (W) dimensions. To ensure effective Sim-to-Real translation without prior real-world training, the network outputs spatial coordinates of the intended grasp target within the observed image, represented by a bounding box (bounding box) notation. The output format is denoted as a vector
\[a = [x_l^{\min}, y_l^{\min}, x_l^{\max}, y_l^{\max}, x_g^{\min}, y_g^{\min}, x_g^{\max}, y_g^{\max}] \in \mathbb{R}^8\]

within the \(R^8\) space, captures the target's bounding box coordinates on both local (\(l\)) and global (\(g\)) images. This output subsequently informs the robot's grasping actions through a closed-loop mechanism based on the computed bounding box predictions.

The architecture of the grasping agent is partitioned into two principal components. First, the YOLOv8 model serves as the object detector. For efficient adaptation from pre-existing models, input resolutions are standardized to 512 \(\times\) 512 pixels. The final segment involves a robot control unit that orchestrates the actual grasping maneuvers and robot movements, grounded on the predicted bounding box coordinates and the concurrent visual inputs. The control unit's objective is to minimize grasping inaccuracies through an error function, 
\begin{equation}
    Error = \alpha \cdot (y_l^{\text{pred}} - y_l^{\text{gt}}) + (1 - \alpha) \cdot (y_g^{\text{pred}} - y_g^{\text{gt}}),
\end{equation} 
where \(y_*^{\text{pred}}\) and \(y_*^{\text{gt}}\) represent the centers of the predicted and actual (ground truth) bounding box, respectively, across both camera views. The system aims for the predicted target's bounding box to coincide with an actual graspable location, effectuating a successful grasping action as the error approaches zero.

\begin{figure}
\centering
  \includegraphics[width=8.7cm]{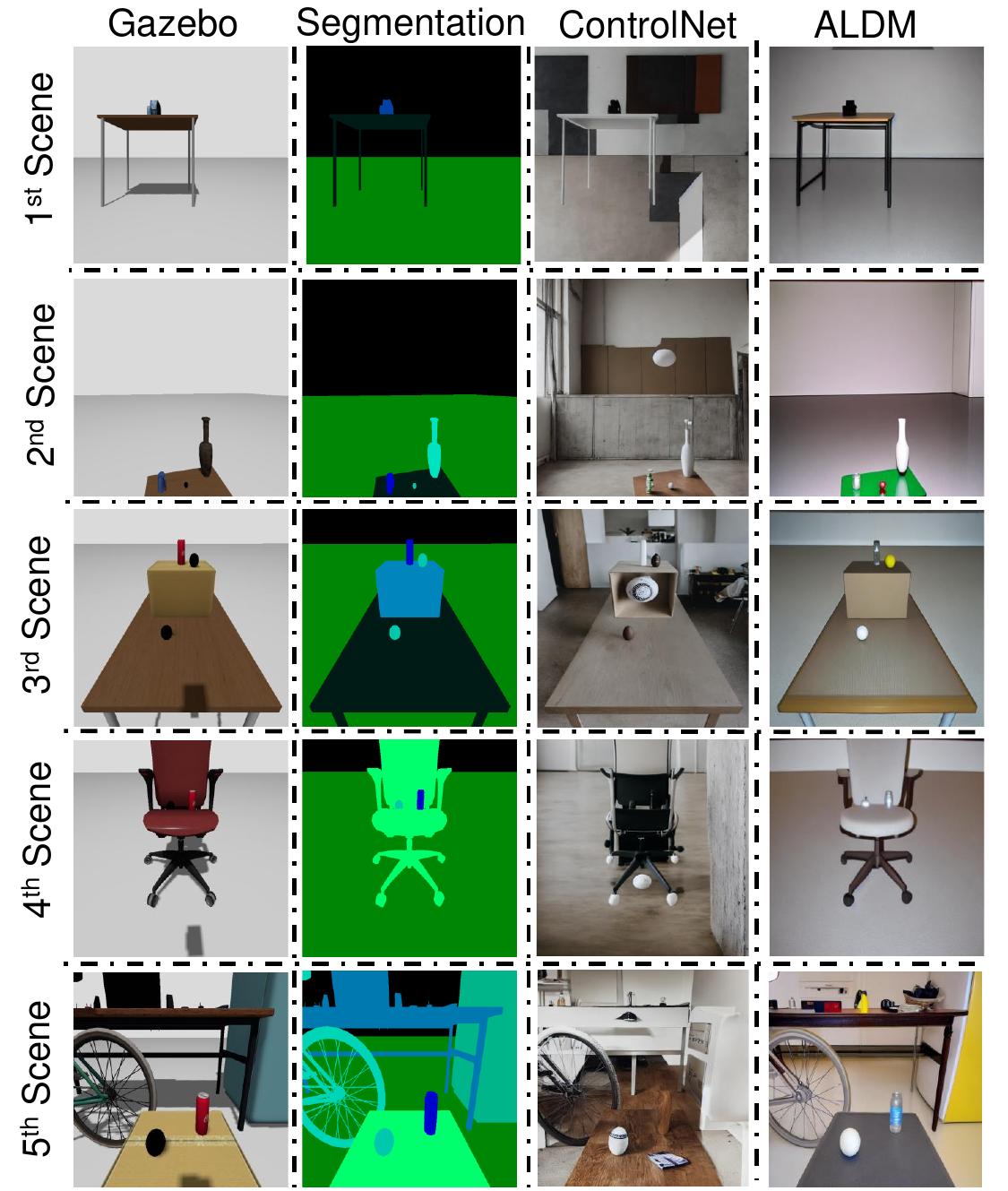}
\caption{Experiment results in image generation. In the \textbf{Gazebo column}, we constructed five simulation scenarios in the Gazebo framework, meticulously designed to align with the physical layout of the laboratory environment. In the \textbf{Segmentation column}, the outcomes of object segmentation, derived from the simulated settings, serve as the inputs for the layout-to-diffusion model, delineated through predefined labels. In the \textbf{ControlNet column}, the images show incorrect image contents and inaccurate object positions. In the \textbf{ALDM column}, it shows the output generated by the image synthesis model. Notably, the produced imagery not only exhibits precise localization of the target objects but also embodies a stylistic resemblance to the actual environmental setting, demonstrating the model's efficacy in bridging the gap between the simulated images' style and laboratory environment.}
\label{fig:result1}       % Give a unique label
\end{figure}

\begin{figure*}
\centering
  \includegraphics[width=\linewidth]{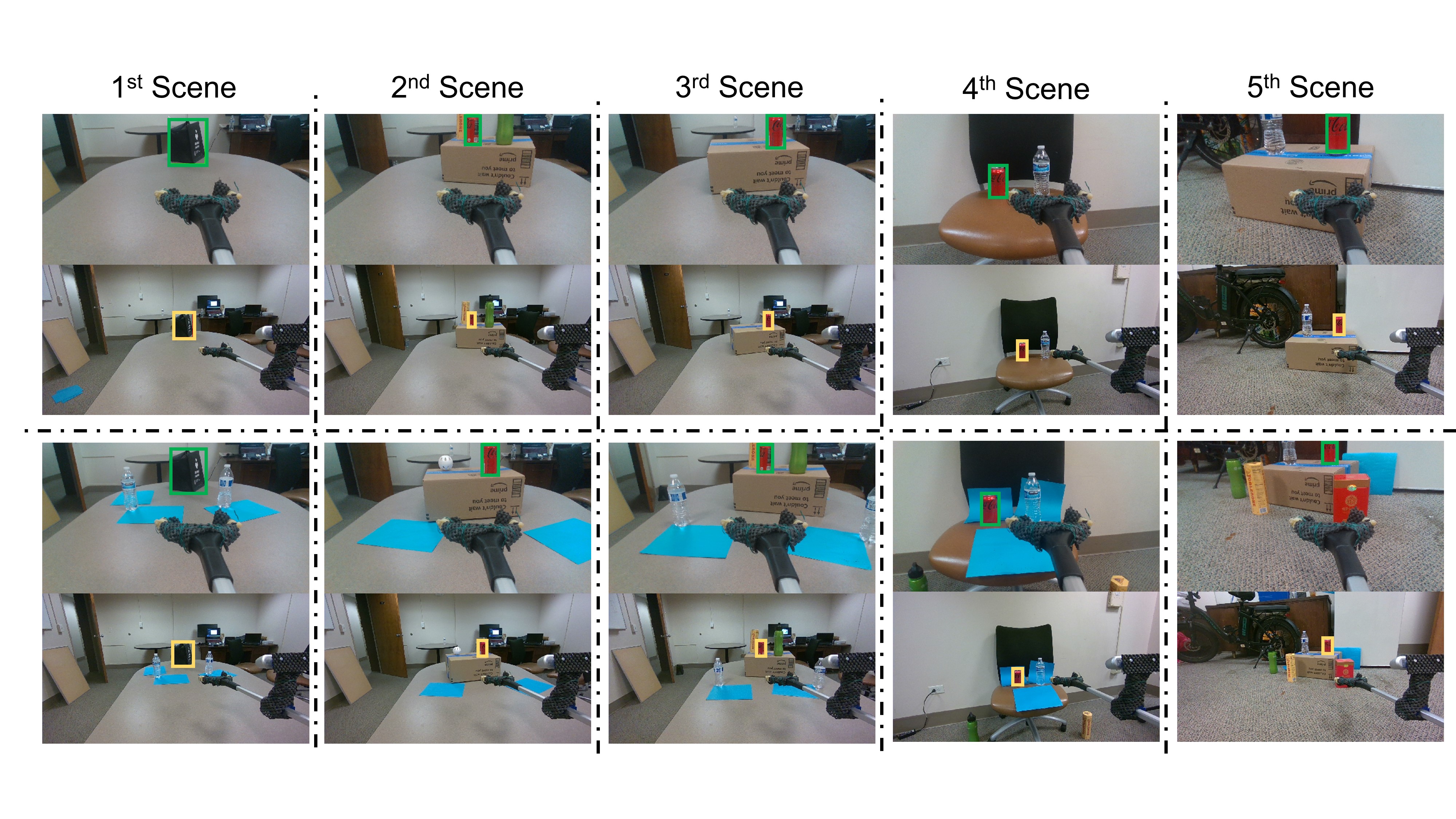}
\caption{The object detection results of the physical robot. The first panel represents the detection performance of the simple background, and the second panel represents the complex background. It can be seen that the target objects are successfully detected while the background has lots of interference objects.}
\label{fig:result2}       % Give a unique label
\end{figure*}

\begin{table*}[ht]
\caption{Comparison of sim-to-real transformation quality across different methods. Appearance fidelity is evaluated using the Inception Score (IS), while Object Detection with YOLOv8 is assessed using typical metrics (Precision-P, Recall-R, mAP50, and mAP50-90).}
\label{table-1-sim2real-result}
\begin{center}
\begin{tabular}{@{}lccccccc@{}}
\toprule
\multirow{2}{*}{Method} &
\multicolumn{1}{c}{Appearance Fidelity} & 
\multicolumn{6}{c}{Object Detection}\\
\cmidrule{2-3}
\cmidrule{5-8}
& IS $\uparrow$ & & & P & R & mAP50 & mAP50-95\\
\midrule
Sim-Only    & - & & & 0.421 & 0.353 & 0.460 & 0.255 \\
CycleGAN~\cite{zhu2017unpaired}    & 3.656 & & & 0.575 & 0.444 & 0.554 & 0.284 \\
ControlNet~\cite{zhang2023adding}   & 5.516 & & & 0.331 & 0.560 & 0.499 & 0.279 \\
ALDM~\cite{li2024adversarial}      & \textbf{5.641} & & & 0.538 & 0.574 & \textbf{0.633} & \textbf{0.313} \\ 
\bottomrule
\end{tabular}
\end{center}
\end{table*}

\section{EVALUATIONS}
\subsection{Datasets and Simulation Environment}
\subsubsection{The training datasets of the image generation model}

To make the ALDM model obtain the knowledge of real-world features and have the ability to generate realistic style images, we have used two public datasets to train the model, which is:
    \begin{itemize}
        \item The \textbf{ADE20K }dataset~\cite{zhou2017scene}, jointly developed by the Massachusetts Institute of Technology and the University of Toronto, comprises over 22,000 images with detailed pixel-level annotations across a wide range of scenes and objects. It is particularly valuable for computer vision tasks that demand a comprehensive understanding of both the broader scene and the finer details of objects, including image segmentation and scene analysis. The dataset's extensive annotations, which cover a broad spectrum of objects and their parts with precision, make it an essential tool for researchers aiming to advance the field of detailed scene interpretation. 
        \item  The \textbf{Cityscapes} dataset~\cite{cordts2016cityscapes} offers a comprehensive collection of over 25,000 urban scene images, designed to enhance semantic understanding of urban street environments. It uniquely combines 5,000 pixel-level annotated images for detailed analysis and 20,000 coarsely annotated images to support a broad range of vision tasks, notably in autonomous driving technologies. Captured across diverse cities, Cityscapes contains a variety of urban landscapes, seasonal variations, and weather conditions, making it a robust dataset for training and evaluating semantic segmentation models. This dataset meticulously annotates numerous elements, including vehicles, pedestrians, and urban fixtures, making it a useful tool for researchers working on innovative ways to analyze urban scenes and improve how intelligent systems understand these environments.
    \end{itemize}
    
These two public datasets offer the model the ability to generate realistic-style images from diverse robot application environments, such as indoor bedrooms, indoor kitchens, outdoor scenes, agricultural scenes, autonomous driving, etc.  

\subsubsection{Experimental Dataset}
% To testify performance of the image generation models, a reliable validation dataset is needed.
All the simulation scenes are built on the Gazebo platform. Examples can be seen in Figure \ref{fig:result1}, five independent scenarios were built, with the semantic camera set to record the segmentation image of different angles. Because the objects in the simulation environments are labeled, the segmentation masks or a bounding box can be easily obtained and work as the ground-truth label dataset in this experiment. In this research, 1235 experimental pairs were created. Each pair has a simulation image (Figure \ref{fig:result1} Gazebo column), cropped by the simulation camera, an instance segmentation image, and a semantic segmentation image (Figure \ref{fig:result1} Segmentation column).

\subsection{Experiments in Image Generation}
The key objective is to guarantee that the model can generate high-quality images that are suitable for training the object detection network (YOLOv8), thus the most important evaluation metric of this part is the deviations between the centers of ground-truth bounding boxes and the bounding boxes detected by the object detection model. 

To obtain the training datasets in the comparison experiments, the segmentation images are sent to the trained CycleGAN, the fine-tuned ControlNet(stable-diffusion-v2.1\cite{Rombach_2022_CVPR}), and ALDM(These three models were both trained on ADE20k to compare the performance) to generate realistic-style images. The images generated by these three models were detected by YOLOv8 separately and analyzed quantitatively.

Table \ref{table-1-sim2real-result} quantitatively evaluates the quality of sim-to-real transfer. In the Appearance Fidelity section, we utilize the Inception Score (IS) to determine whether the sim-to-real generated images possess sufficient fidelity. In the Object Detection section, considering that images transferred from sim-to-real with superior spatial consistency are more effective as training data for detection, we incorporate sim-to-real images generated by different methods into the YOLOv8 detection training dataset and evaluate them using the same real-image test set. As shown in Table \ref{table-1-sim2real-result}, ALDM outperforms in both aspects, showcasing its exceptional zero-shot sim-to-real transfer capabilities. Interestingly, although the ControlNet model generates images of higher fidelity compared to CycleGAN, its detection outcomes are inferior. This may be attributed to ControlNet's emphasis on appearance while neglecting spatial consistency.

\begin{table}[t!]
\centering
\caption{The success rates of grasping in 20 trials for different methods with two types of backgrounds}
\label{table-2}
\begin{tabular}{@{}lcc@{}}
\toprule
Methods          & Plain background & Complex background \\ \midrule
Sim-Only         & 10\%              & 0\%                \\
CycleGAN       & 0\%              & 0\%                \\
ControlNet   & 25\%             & 5\%               \\
ALDM    & 75\%             & 65\%               \\ \bottomrule
\end{tabular}
\end{table}

\subsection{Experiments in Real Environment With Zero-Shot Sim-to-Real Transfer}
The experiment assessed the grasping success rate across five real-world environments, ranging from basic to complex setups augmented with novel visual noise to challenge the segmentation model. As detailed in Table \ref{table-2}, the experiment's results highlight the varying success rates of object grasping methods in 20 trials across plain and complex backgrounds. The failure of CycleGAN is attributed to its inability to generalize to complex testing environments when faced with varying backgrounds. ControlNet's underperformance is due to inadequate object generation quality and quantity. Conversely, our ALDM-based model excels not only in simple backgrounds, achieving a 75\% success rate in object grasping tasks but also maintains a considerable 65\% success rate in complex environments. This robust performance of ALDM can be credited to its capacity to generate high-quality images and the robustness of the algorithm, which stands as a testament to its superiority in handling object-grasping tasks under diverse conditions. 

\section{CONCLUSION \& DISCUSSION}

This paper presents a groundbreaking framework, ALDM-Grasping, that utilizes ALDM for zero-shot Sim-to-Real transfer in visual grasping tasks. The model excels in generating controlled image content from textual descriptions, pinpointing object grasp positions, and demonstrating zero-shot learning in complex, unseen scenarios.
Future work will extend ALDM to diverse gripper configurations and explore its applicability in 3D unstructured settings, such as robotic fruit harvesting and autonomous driving. Additionally, we aim to test ALDM's capabilities in varied manipulation tasks, including rotation and placement, to assess its versatility. The exploration of more vision backbone models will further refine ALDM's effectiveness in Sim-to-Real applications.

\bibliographystyle{IEEEtran} 
\bibliography{ref.bib}

\end{document}